\documentclass[3p,12pt]{elsarticle}
\usepackage{epsfig}
\usepackage{subfigure}
\usepackage{textcomp}
\usepackage{float}
\usepackage{CJK,CJKnumb,CJKulem}
\usepackage{rotating}
\usepackage{mdwlist}
\usepackage{amssymb}
\usepackage{amsfonts}
\usepackage{url}
\usepackage{algorithm}
\usepackage{algorithmic}
\usepackage{multirow}
\usepackage{amsmath}
\usepackage{xcolor}
\usepackage{lineno,hyperref}
\usepackage{graphicx}
\usepackage{epstopdf}
\usepackage{bm}
\usepackage{array}
\usepackage{wrapfig}
\newcommand{\PreserveBackslash}[1]{\let\temp=\\#1\let\\=\temp}
\newcolumntype{C}[1]{>{\PreserveBackslash\centering}p{#1}}
\newcolumntype{R}[1]{>{\PreserveBackslash\raggedleft}p{#1}}
\newcolumntype{L}[1]{>{\PreserveBackslash\raggedright}p{#1}}

\modulolinenumbers[5]

\begin{document}
\begin{frontmatter}

\title{Learning Efficient Convolutional Networks through Irregular Convolutional Kernels}


\author[myaddress,coaddress]{Weiyu Guo}

\author[coaddress]{Jiabin Ma}

\author[coaddress]{Liang Wang}

\author[wataddress,coaddress]{Yongzhen Huang}
\address[myaddress]{Information School, Central University of Finance and Economics, Beijing, P.R.China,  102206}
\address[coaddress]{National Laboratory of Pattern Recognition, Institute of Automation,\\ Chinese Academy of Sciences, Beijing, P.R.China, 100190}
\address[wataddress]{Watrix technology limited co. ltd, China\\
Email: weiyu.guo@cufe.edu.cn, jiabin.ma@cripac.ia.ac.cn, wangliang@nlpr.ia.ac.cn, hyz@watrix.ai}


\begin{abstract}
As deep neural networks are increasingly used in applications suited for low-power devices, a fundamental dilemma becomes apparent: the trend is to grow models to absorb increasing data that gives rise to memory intensive; however low-power devices are designed with very limited memory that can not store large models. Parameters pruning is critical for deep model deployment on low-power devices. Existing efforts mainly focus on designing highly efficient structures or pruning redundant connections for networks. They are usually sensitive to the tasks or relay on dedicated and expensive hashing storage strategies. In this work, we introduce a novel approach for achieving a lightweight model from the views of reconstructing the structure of convolutional kernels and efficient storage. Our approach transforms a traditional square convolution kernel to line segments, and automatically learn a proper strategy for equipping these line segments to model diverse features. The experimental results indicate that our approach can massively reduce the number of parameters (pruned 69\% on DenseNet-40) and calculations (pruned 59\% on DenseNet-40) while maintaining acceptable performance (only lose less than 2\% accuracy).
\end{abstract}

\end{frontmatter}
\section{Introduction}
With the rapid development of the deep learning, Convolutional Neural Networks (CNNs)\cite{cnnLecun1989backpropagation} have become ubiquitous ranging from image classification\cite{lecun1998gradient,krizhevsky2012imagenet} to semantic segmentation \cite{long2015fully,chen2016deeplab} and object detection\cite{girshick2015fast,ren2015faster,liu2016ssd}, since them effectively extracts valuable and abstract features. Although deep models \cite{krizhevsky2012imagenet,simonyan2014very,szegedy2016rethinking,he2016deep, huang2016densely} are very powerful, the large number of learnable parameters leads to a mass of calculations and memory of devices consumption. For example, the parameter numbers of ResNet101\cite{he2016deep} and DenseNet100\cite{huang2016densely} are 39M and 27M, respectively. It leads to many of low-power devices are hard to deploy CNNs. Evidently, deep neural networks would be used more widely if their computational cost and storage requirement could be significantly reduced.
\begin{figure}[!t]
\centering
\includegraphics[height=0.25\linewidth,width=0.7\linewidth]{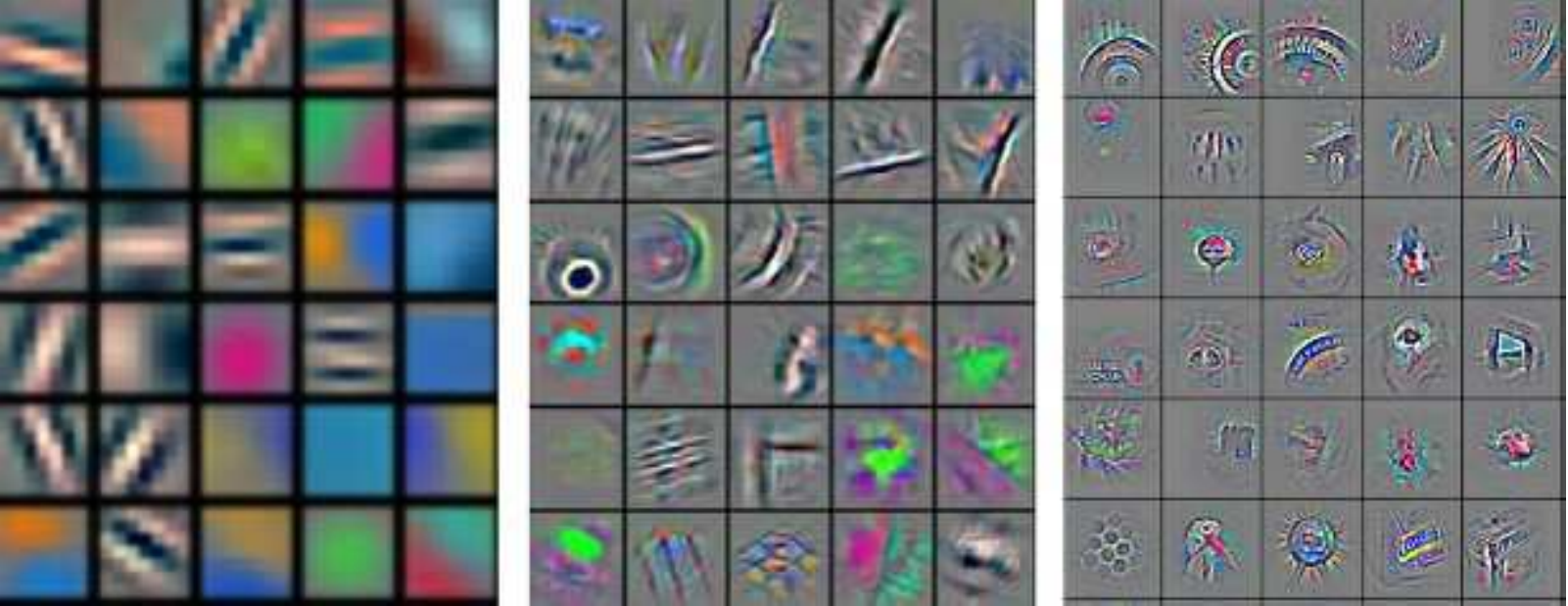}
\caption{Feature maps generated by AlexNet from low to high convolution layers.}
\label{fig1}
\end{figure}

Recently, an increasing line of effort has been undertaken to tackled compressing the size of models for CNNs. Most of existing approaches are considering to design delicate structures or prune redundant connections of networks. Kernel factorization and weight pruning are two branches of representative approaches to reduce the model size of deep neural networks. Kernel factorization like Inception-V3\cite{szegedy2016rethinking} uses asymmetric kernels like ${n\times1}$ and ${1\times{n}}$, and perpendicularly intersects them to replace one normal square kernel. Since asymmetric kernels have line segment shapes, this kind of two-layer solution is ${33\%}$ cheaper than normal square kernel for the same number of output channels. Weight pruning\cite{han2015deep, venkatesh2016accelerating} is a kind of post processing methods. It deletes the connections which weights are smaller than setting thresholds. The thoughts of these two kinds of methods are both to suppose that the generic architecture of CNNs have much parameter redundancy, and exist even smaller kernels than original square kernels are competent for feature extraction. However, existing methods have some limitations. Such methods of asymmetric convolution kernel factorization only have two kinds of angles, i.e., vertical and horizontal, which somehow restricts the model capability of convolution. While the approaches of common weight pruning can not be completed in one step as they need to wait for the whole kernels having been pre-trained and then prune and fine-tune them. Moreover, common weight pruning models neither reduce computing cost nor speed up original models due to their pruned models needing to be saved by complicated hashing techniques.
\begin{figure}[!t]
\begin{center}
\begin{minipage}[t]{0.5\linewidth}
\centerline{\includegraphics[height=0.5\linewidth,width=0.93\linewidth]{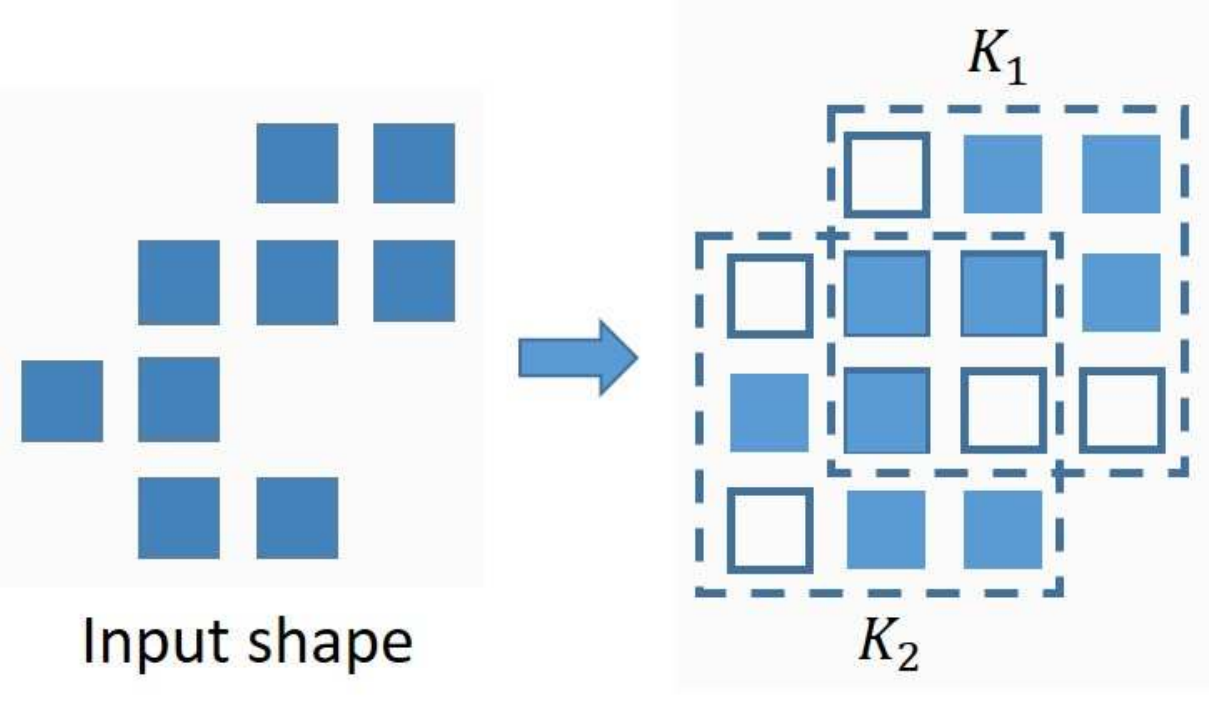}}
\centerline{(a)}\medskip
\end{minipage}

\begin{minipage}[t]{0.4\linewidth}
\centerline{\includegraphics[height=0.45\linewidth,width=0.5\linewidth]{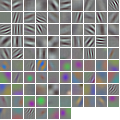}}
\centerline{(b)}\medskip
\end{minipage}
\hfill
\begin{minipage}[t]{0.4\linewidth}
\centerline{\includegraphics[height=0.45\linewidth,width=0.96\linewidth]{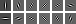}}
\centerline{(c)}\medskip
\end{minipage}
\vfill
\caption{(a) Multiple kernels assembly. (b) Feature maps visualization of AlexNet. (c) Kernels visualization of Gabor.}
\end{center}
\label{fig2}
\end{figure}

In this work, we aim to significantly prune the number of parameters for CNN models without additional computing cost caused by complicated hashing techniques, as well as maintain acceptable performance. To achieve these targets, there are several challenges we have to face. First, the redundant parameters in CNN models usually take on stochastic distribution. Removing such redundant parameters will result in sparse matrices of parameters, which leads to additional cost for hashing in phases of saving and loading models. Second, most of accelerated libraries for deep neural networks, e.g, CuDNN\cite{cudnn} and OpenBlas\cite{openblas}, are developed for coping with dense matrix. It a challenging task to leverage accelerated libraries to speed up the pruned networks which weights matrices are sparse. Finally, as Figure~\ref{fig1} shown, the key of CNN models that distinguish them from other models in computer vision is the convolutional operation, like 3$\times$3 convolution, which can effectively extract and assembly edges, angles and shapes from low to high features by learning the assembly patterns of convolutional kernel layer by layer. Intuitively, we can abstract the ``principal components'' from convolutional kernels to achieve an efficient assembly. As shown in Figure~2(a), an irregular shape can be modeled by assembling two ``principal components'' of squares. However, like Figure~2(b), feature maps are usually various. Thus, as shown in Figure~2(c) the trained convolutional kernels should have various ``principal components'', which causes lots of redundant parameters. It is very challenging to maintain powerful abilities of CNN kernels to extract and assembly features, as well as, find a general pattern to efficiently express the ``principal components''.
%
\begin{figure}[!t]
\centering
\includegraphics[height=0.3\linewidth,width=0.7\linewidth]{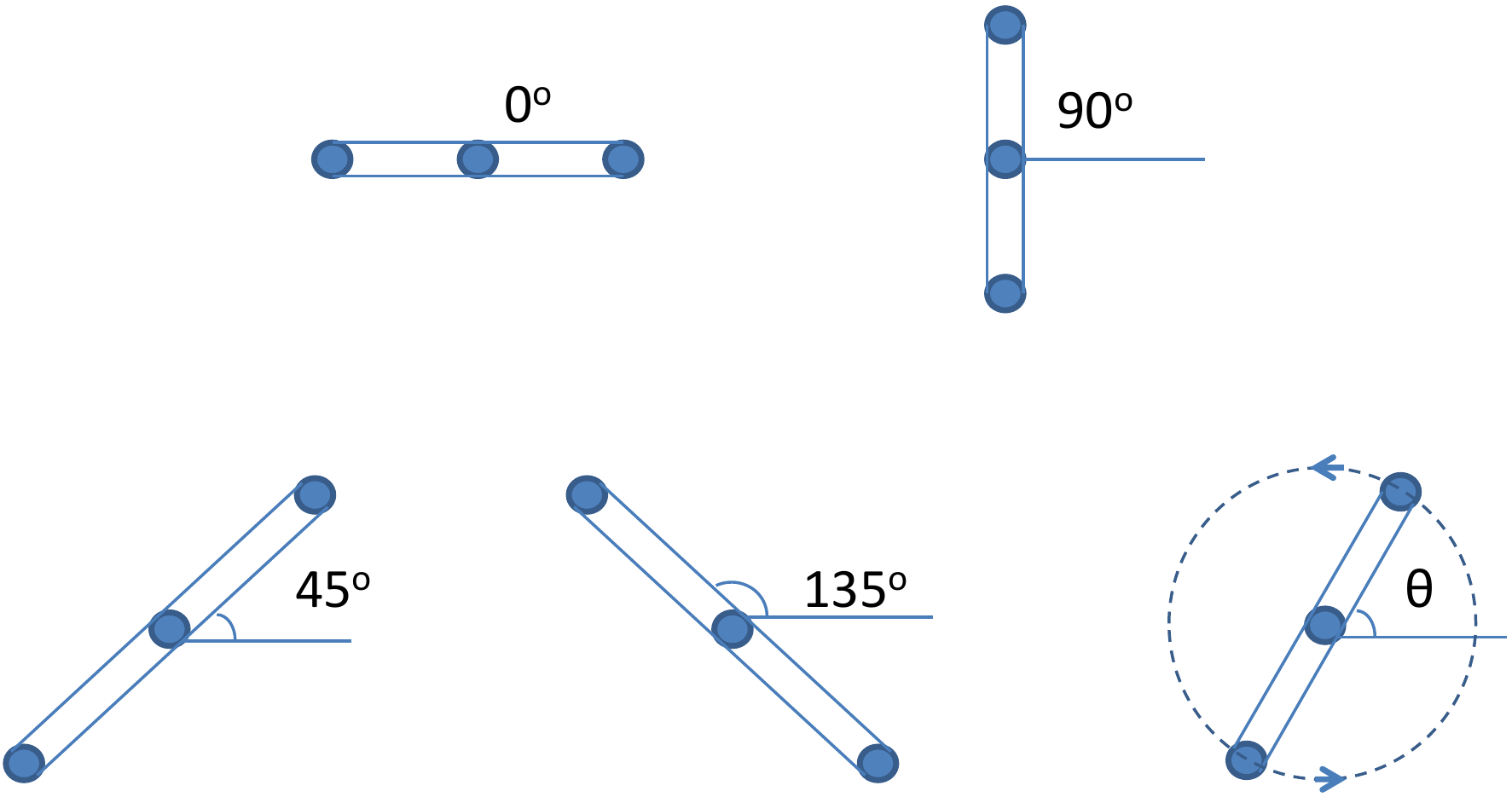}
\caption{Asymmetric kernels with different angles. In top-down and left-right order, the former four kernels have different but fixed angles as ${\rm{0}}^\circ$, ${\rm{90}}^\circ$, ${\rm{45}}^\circ$ and ${\rm{135}}^\circ$. The last one is an illustration of our RotateConv kernel that is asymmetric but rotatable, which means the angle $\theta$ is learnable during training.}
\label{angles}
\end{figure}
To address these problems, we propose a simple, efficient, yet effective method, known as Rotated Convolution(RotateConv), to makes up the deficiency of traditional kernel factorization and weight pruning. Traditional convolutional kernels, which shapes are usually square or rectangle, can learn the patterns in feature maps by overlapped scanning. As shown in Figure~2(c), AlexNet has learned a variety of frequency- and orientation-selective kernels. These kernels have clear skeletons which are composed by ``black grids'', and most of them present as line segments. Inspired by such evidence, Our RotateConv is shown as the last one in Figure~\ref{angles}. Aiming at assembling ``principal components'', the basic shape of RotateConv kernel is a line segment, which means that it only has 2 or 3 weights for convolution. Besides weights, a RotateConv kernel has an additional learnable variable, namely, angle $\theta$, which makes RotateConv is much more flexible considering that directions from angles ${\rm{0}}^\circ$ to ${\rm{180}}^\circ$ instead of ${\rm{0}}^\circ$ and ${\rm{90}}^\circ$ than asymmetric kernels of kernel factorization. Note that making the kernel rotatable is achieved by making the variable $\theta$ continuous and learnable. Due to the existence of $\theta$, these line segment kernels can have a larger receptive field like ${3\times3}$, so its modeling capacity still can be guaranteed. In addition, for avoiding the computing cost caused by complicated hashing techniques in the phase of loading pruned model, we leverage interpolation-based algorithms to realize efficiently storing and restoring pruned model instead of hashing techniques. Considering different emphases on the model size and performance, we propose two variants of rotatable kernels. One has $n+1$ parameters consisting of $n$ weights and an angle value for each convolutional kernel. And the other one only has $n$ parameters for each convolutional kernel and one additional parameter for each convolutional filter. Accordingly the compress ratios are ${(n+1)/9}$ and approximately equal to ${n/9}$ for $3\times3$ convolution, respectively. 


Compared with the existing work, our contributions in this work can be summarised as follows.
\begin{itemize}
\item Compared with asymmetric kernels of kernel factorization, proposed RotateConv kernels are much more flexible considering that they can have any continuous angles from ${\rm{0}}^\circ$ to ${\rm{180}}^\circ$ instead of ${\rm{0}}^\circ$ and ${\rm{90}}^\circ$. Compared with previous methods of weight pruning, our RotateConv kernels do not need to pretrain, as they not only have much less weights than the square one, but also are born with line segment shape that can be trained end to end.\vspace{-0.5em}

\item Compared with previous methods which store and restore pruned models by hashing techniques, we propose learnable and interpolation-based methods to efficiently store and restore pruned model. Taking this advantage, our pruned CNN models have enormous potential to run faster on low-power devices, such as ARM CPU and FPGA.
\item We conduct a series of experiments to validate the proposed methods. The results show that our final pruned models achieve competitive performance, e.g., for SSD\cite{liu2016ssd}, we can reduce the number of parameters 80\% and save more than 60\% FLOPs without accuracy slumping.
\end{itemize}

\section{Related Work}
Since RotateConv devotes to achieve a more compressed convolutional kernel, related work can be divided into two groups, i.e., convolution kernel design and model compression.

\subsection{Convolution kernel design}
The most representative work of kernel design in deep CNN can be inferred from the series work of Inception-V$n$\cite{szegedy2015going,ioffe2015batch,szegedy2016rethinking}. Inception-V1 uses multi-scale kernels to extract scale-invariant features. Inception-V2 uses more stacked small kernels to replace one bigger kernel so as to increase the depth of the network and reduce the number of parameters. Inception-V3 further makes the kernel smaller as it uses asymmetric kernels like ${1\times3}$ and ${3\times1}$. As we mentioned above, though this asymmetric kernel reduces parameters a lot, its fixed angle as vertical and horizontal puts limitations on the capacity of modeling more orientation-flexible patterns. Dilated convolution\cite{yu2015multi} is another widely used method for convolution kernel design, which aims at solving the resolution reduction problem of feature maps in forward propagation. It is a dilated variant of traditional compact kernels, which helps the kernel have a larger receptive field without increasing parameters. However, dilated kernels usually result in memory cache missing problem and suffer from unexpected speed bottleneck.


Recently, there emerge some novel deformable kernel design works. Deformable Convolutional Networks (DCN)\cite{dai2017deformable} is a recently proposed excellent work which learns irregular kernels. DCN has a similar thought with Region Proposal Network\cite{ren2015faster} as it applies a usual convolution on the input feature, and then outputs the new kernel shape for the following deformable convolution layer. Irregular Convolutional Neural Networks (ICNN)\cite{ma2017irregular} is another work learning irregular kernels. Different from DCN, ICNN directly models the kernel's shape attributes as learnable variables and learns the shapes in the same way as kernel weights. Although, these two methods can expand the capacity of convolutional kernels, they utilize extra parameters and made the calculation more complicated. Unlike existing methods, our methods change the shape of the convolutional kernels for devoting to maintain the capacity of CNN models with less parameters.

\subsection{Model compression}
There has been growing interest in model compression due to the demands of device-limited applications. From the viewpoint of having potential to reduce the requirement of storage and accelerate the deep models in the phase of inference, the related work of model compression can be ranged into low-rank decomposition, weight sparsifying, structured pruning and quantization representation.

Low-rank decomposition methods approximate weight matrix of neural networks with low-rank techniques like Singular Value Decomposition\cite{svd} or CP-decomposition\cite{cpDecomposition}. These methods usually can obtain well performance on ``big layers'', such as fully-connected layers and such convolution layers using big kernels like 5$\times$5 and 7$\times$7. Thus, they can yield significant model compression and acceleration on Alexnet\cite{alexnet} with a little compromise on accuracy. However, these methods can not achieve notable effect of compression on modern CNN architectures, since their convolutional layers tend to use small kernels like 3$\times$3 and 1$\times$1 kernels.

Weight sparsifying methods attempt to prune the unimportant connections in pre-trained neural networks. The resulting network's weights are mostly zeros thus the storage space can be reduced by storing the model with sparse formats. Song Han et al.\cite{han2015deep} alternately prune the unimportant connections with given thresholds and fine-tune the pruned networks to reduce the number of parameters by 9$\times$ and 13$\times$ for AlexNet and VGG-16 model, respectively. However, this kind of methods only can achieve speedup with dedicated sparse matrix operation libraries and/or hardware, since the sparse weights need to auxiliary operation to retrieval in the phase of inference. Srinivas et al.\cite{Srinivas2017Training} overcome the limitation of generating sparse weights by setting thresholds, they explicitly impose sparse constraints over each weight tensor, and achieve high compression rates with a more efficient training. However, this method still suffers from the same drawback with Song Han's work \cite{han2015deep}, i.e., they are easy to obtain small models, but hard to really speed up networks with general computing devices like CPU, since the work of processing sparse matrix is not friendly for CPU.

Structured pruning works are proposed for pruning redundant structures, such as channels and layers, in trained deep models. Channel pruning methods\cite{pruningFilters,learningEfficient} are most representative work in this branch. Li et al.\cite{pruningFilters} prune input channels for each layer indicating by the weights of convolutional kernels, while Liu et al.\cite{learningEfficient} learn a scaling factor which indicates the importance of channel for each output channel, then heuristically prunes the output channels with learned scaling factors. Changpinyo et al.\cite{powerSparsity} introduce sparsity both on input and output channels by random deactivating input-output channel-wise connections. To achieve a full-scale solution, Wen et al.\cite{structuredSparsity} utilize a Structured Sparsity Learning (SSL) method to sparsify different level of structures (e.g. filters, channels or layers) in CNNs. He et al.\cite{channelPruning} effectively prunes each layer by channel selection based on LASSO regression\cite{hans2009bayesian} and least square reconstruction of output. All of these methods utilize sparsity regualarization during training to obtain structured sparsity. Compared with weight sparsifying, these channel pruning works overcome the limitation of requiring to restore compact matrices or dedicated speeding up libraries.

Quantization representation methods quantize network weights from float(usually 32 bits) to be a few of bits. Binary-net\cite{binaryNet} and XNOR-net\cite{xnorNet} can achieve 32$\times$ compression rates and 58$\times$ speed up on deep models by using just one bit to store one weight and bitwise operations, but often notably sacrifice accuracy. HashNet\cite{hashNet} quantizes the network weights by a hash strategy, while Song Han et al.\cite{han2015deep} quantizes the network weights by a clustering strategy. These methods assign the network weights to different groups and within each group weight the value is shared. In this way, only the shared weights and indices of weights need to be stored, thus a large amount of storage space could be saved. However, these techniques can neither save run-time memory nor inference time unless they are aided by special devices, since the shared weights need to be restored to their original positions before calculation.

\section{Mathematical Derivation of Rotated Convolution}\label{rck}
Early work\cite{han2015deep,hashNet,channelPruning} have proved that only reserving the ``principal components'' of CNN models, i.e., a few of importance weights, the CNN models still can maintain their ability of feature extraction. In this section, we aim to prune and compress standard convolution networks to such ``principal components'' by elaborating Rotated Convolution Network (RCN). Accordingly, we first formulate our rotated convolution kernel for RCN, and present in detail how we efficiently achieve RCN which has both high compression rate and accuracy from standard CNNs. Then we will explain RotateConv's mathematical derivations.

\subsection{Rotated convolution kernel}
\begin{figure}[t]
\begin{center}
\hspace{15pt}
\begin{minipage}[t]{0.275\linewidth}
  \centering
  \centerline{\includegraphics[height=0.6\linewidth,width=0.6\linewidth]{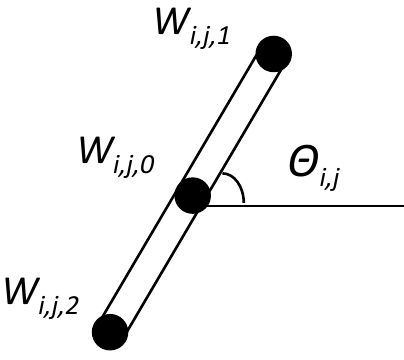}}
  \centerline{(a). Theoretical kernel}\medskip
\end{minipage}
\begin{minipage}[t]{0.275\linewidth}
  \centering
  \centerline{\includegraphics[height=0.6\linewidth,width=0.6\linewidth]{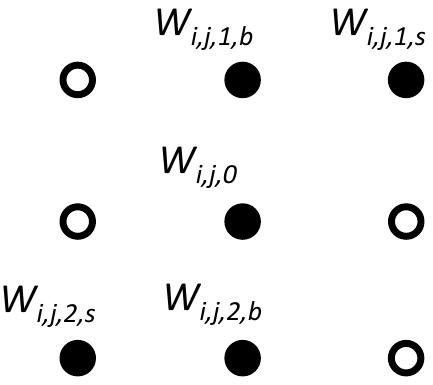}}
  \centerline{(b). Practical kernel}\medskip
\end{minipage}
\begin{minipage}[t]{0.35\linewidth}
  \centering
  \centerline{\includegraphics[height=0.6\linewidth,width=0.6\linewidth]{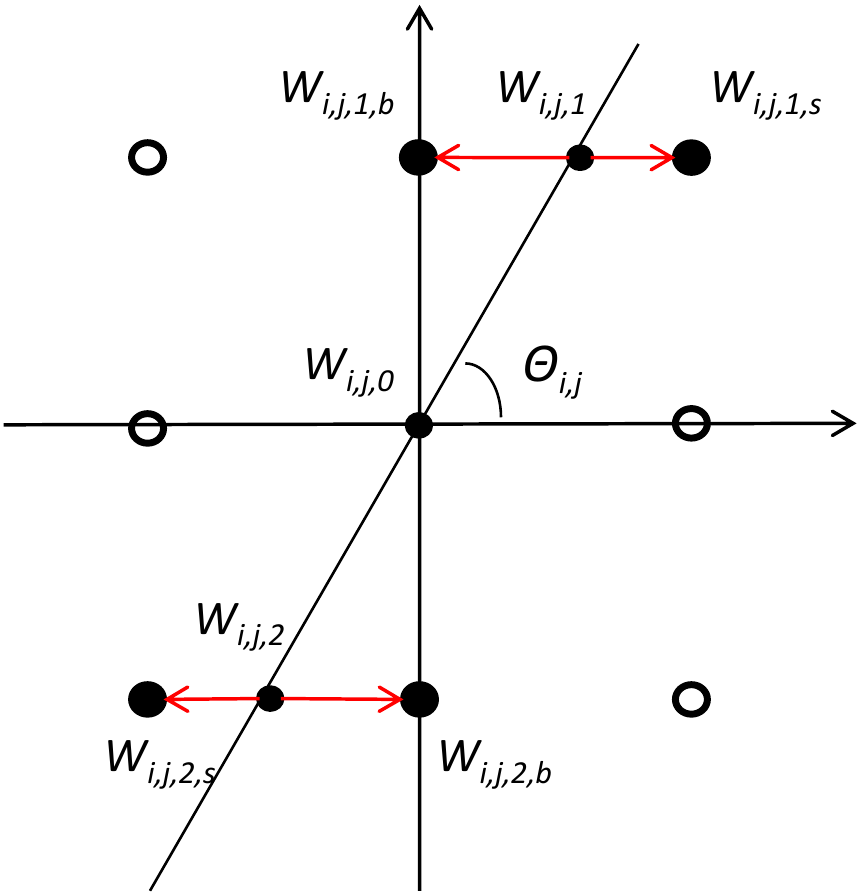}}
  \centerline{(c). Angle inverse-interpolation}\medskip
\end{minipage}
\end{center}
\caption{(a) Theoretical RotateConv kernel has a line segment shape, whose weights are ${w_0}$, ${w_1}$, ${w_2}$ and angle is ${\theta}$. (b) Practical RotateConv kernel for computation after inverse-interpolation based on angle. (c) The inverse-interpolation process for RotateConv.}
\label{Figure3}
\end{figure}

Distinguish from a traditional CNN model which convolutional kernels have fixed shapes and can be expressed as matrices, e.g., a standard 3$\times$3 kernel is expressed as a 3$\times$3 matrix, our RCN model is formed by rotated convolution kernels which consists of 3 weights placed on a straight line as shown in Figure~\ref{angles}. Accordingly, instead of using a matrix, we formulate the rotated convolution kernel $\widetilde{K}$ by building coordinate systems on its corresponding standard kernel $K$.
\begin{equation}\small
\begin{split}
&\widetilde{K} = \{ W,T\} \\
&W = \{ {w_{i,j,0}},{w_{i,j,1}},{w_{i,j,2}}|i = 1,2,...,N,j = 1,2,..,M\} \\
&T = \{ {\theta _{i,j}}|i = 1,2,...,N,j = 1,2,..,M\}
\end{split}
\end{equation}
where ${M}$ denotes the number of input channels and ${N}$ is the number of output channels. ${W}$ is the set of kernel weights and each kernel has 3 weights. ${T}$ is the set of kernel angles $\theta$. $\theta$ is defined as the included angle between the horizontal line and the kernel line as shown in Figure~\ref{angles}, which is range in  ${\rm{0}}^\circ$ to ${\rm{180}}^\circ$. For a RotateConv kernel shown in Figure~\ref{Figure3}(a), the output can be calculated as:
\begin{equation}
{S_i} = \sum\limits_{j = 1}^{M} {\left( {{w_{i,j,0}}{I_{j,0}} + {w_{i,j,1}}{I_{j,1}} + {w_{i,j,2}}{I_{j,2}}} \right)}
\end{equation}
where ${S_i}$ is the weighted summation in the ${i}$-th output channel. Inputs ${I_{j,0}}$, ${I_{j,1}}$ and ${I_{j,2}}$ correspond to the weights ${w_{i,j,0}}$, ${w_{i,j,1}}$ and ${w_{i,j,2}}$, respectively. Note that for RotateConv, ${I_{j,1}}$, ${I_{j,2}}$ are sampled by ${\theta_{i,j}}$.

\subsection{Interpolation based on angle}
The rotated convolution kernel $\widetilde{K}$ can be viewed as the ``principal components'' of $K$. Hence the calculation of convolution only needs to process the pixels corresponding to ``principal components''. However, as ${\theta_{i,j}}$ in $T$ is defined to be continuous, if ${\theta_{i,j}}$ does not equal to an integer multiple of ${\rm{45}}^\circ$, then it does not exist corresponding pixels, i.e., ${I_{j,1}}$ and ${I_{j,2}}$, on feature maps.

To solve this problem, we first build a coordinate system on its related kernel $K$, which is shown as Figure~\ref{Figure3}(b). Then, we have two choices to match input pixels with weights. Specifically, we can combine the input pixels corresponding to ${w_{i,j,1,b}}$ and ${w_{i,j,1,s}}$ or split the weights ${w_{i,j,1}}$ and ${w_{i,j,2}}$, to the adjacent positions which are the integer multiple of ${\rm{45}}^\circ$ as shown in Figure~\ref{Figure3}(c). Most of deep learning software frameworks utilize the algorithm of ${im2col}$ to transform the calculation of convolution to be dense matrix calculation. If we adopt combining the input pixels, the ${im2col}$ will invalidate because $\theta$ is a variable. Therefore, in this work, we adopt the second method to avoid extra time consuming related with the practical convolution implementation. And at last, a RotateConv kernel associated with a 3$\times$3 square kernel is equivalent to keeping 5 weights like Figure~\ref{Figure3}(b). The split process can be calculated as:
\begin{equation}\small
\begin{array}{l}
{w_{i,j,1,b}} = {w_{i,j,1}} \times f(\theta_{i,j} )\\
{w_{i,j,1,s}} = {w_{i,j,1}} \times (1-f(\theta_{i,j} ))\\
{w_{i,j,2,b}} = {w_{i,j,2}} \times f(\theta_{i,j} )\\
{w_{i,j,2,s}} = {w_{i,j,2}} \times (1-f(\theta_{i,j} ))\\
\\
s.t.{\rm{  }}f(\theta_{i,j} ) = \frac{{\theta_{i,j} \% 45}}{{45}},{\rm{  0}}^\circ \le \theta_{i,j} {\rm{ < 180}}^\circ
\end{array}
\end{equation}
where ${w_{i,j,1,b}}$ and ${w_{i,j,1,s}}$ mean the weights split by ${w_{i,j,1}}$, same as ${w_{i,j,2,b}}$ and ${w_{i,j,2,s}}$ split by ${w_{i,j,2}}$. The spliting sizes are determined by ${f(\theta_{i,j})}$, which is simply defined as the ratio between the include angle and ${\rm{45}}^\circ$.

To this end, the convolution is a weighted summation after inverse-interpolation based on angle:
\begin{equation}\small
\begin{array}{l}
S_i = \sum\limits_{j = 1}^{{M}} {\left( {{S_{i,j,0}} + {S_{i,j,1}} + {S_{i,j,2}}} \right)} \\
s.t.\left\{ {\begin{array}{*{20}{c}}
\begin{split}
&{{S_{i,j,0}} = {w_{i,j,0}}{I_{i,j,0}}}\\
&{{S_{i,j,1}} = {w_{i,j,1,b}}{I_{i,j,1,b}} + {w_{i,j,1,s}}{I_{i,j,1,s}}}\\
&{{S_{i,j,2}} = {w_{i,j,2,b}}{I_{i,j,2,b}} + {w_{i,j,2,s}}{I_{i,j,2,s}}}
\end{split}
\end{array}} \right.
\end{array}
\end{equation}

\subsection{Back propagation}
There are three kinds of learnable variables, i.e., inputs ${I}$, weights ${W}$ and angles ${T}$, and one kind of intermediate variables ${W_{bs}}$, as called ${w_{i,j,1,b}}$, ${w_{i,j,1,s}}$, ${w_{i,j,2,b}}$ and ${w_{i,j,2,s}}$. The gradients for inputs ${I}$ can be calculated by the same as the ones of traditional convolution, while the gradients for ${W_{bs}}$ and ${W_{i,j,0}}$ also can be calculated with the traditional way, since ${W_{bs}}$, ${W_{i,j,0}}$ and zeros are combined to be a usual ${3\times3}$ kernel as shown in Figure~\ref{Figure3}(b). The key point is to compute the gradients for weights ${W_{i,j,1}}$, ${W_{i,j,2}}$, and angles ${\theta_{i,j}}$ with the intermediate variables ${W_{bs}}$.
\begin{equation}\small
\begin{split}
&\Delta {w_{i,j,1}} = \Delta {w_{i,j,1,b}} f({\theta _{i,j}}) + \Delta {w_{i,j,1,s}}(1 - f({\theta _{i,j}}))\\
&\Delta {w_{i,j,2}} = \Delta {w_{i,j,2,b}} f({\theta _{i,j}}) + \Delta {w_{i,j,2,s}}(1 - f({\theta _{i,j}}))\\
&\Delta {\theta _{i,j}} = {w_{i,j,1}}(\Delta {w_{i,j,1,b}} - \Delta {w_{i,j,1,s}})f'({\theta _{i,j}})\\
&{\qquad\qquad} + {w_{i,j,2}}(\Delta {w_{i,j,2,b}} - \Delta {w_{i,j,2,s}})f'({\theta _{i,j}})
\end{split}
\end{equation}

The update mechanism of ${W_{bs}}$ is the same as before for standard weights ${W}$, but not for angles ${T}$. For a certain angle $\theta$, since backward gradients, are supplied from ${w_{bs}}$, the updated new $\theta$ should not excess the boundaries defined by ${w_{b}}$ and ${w_{s}}$ too much.
\begin{equation}\small
\begin{array}{l}
{\theta _{update}} = ({\theta _{last}} + \Delta \theta )\% 180\\
s.t.{\rm{  }}\left\{ {\begin{array}{*{20}{c}}
\begin{split}
&{{\theta _{last\_small}} - \epsilon < {\theta _{update}} < {\theta _{last\_big}} + \epsilon }\\
&{{\theta _{last\_small}} = {\theta _{last}} - {\theta _{last}}\% 45}\\
&{{\theta _{last\_big}} = {\theta _{last}} - {\theta _{last}}\% 45 + 45}
\end{split}
\end{array}} \right.
\end{array}
\end{equation}
where ${\epsilon}$ is a small positive value to allow ${\theta_{update}}$ to get out of the last adjacent boundaries ${\theta _{last\_small}}$ and ${\theta _{last\_big}}$ but not too much. Note that weights ${W}$ and angles ${T}$ can both be initialized from random distribution or pre-trained ${1\times3}$ and ${3\times1}$ kernels, while the initialization for angles ${T}$ should be in the range from zero to ${\rm{180}}^\circ$.

\section{Arithmetic interpolation}\label{ai}
Although we deduce RotateConv kernels in above section with punning 3$\times$3 convolutional kernels, the punning method can be used to arbitrary scale convolutional kernels. In the meantime, we have noticed that present-day CNN models are constructed by a mass of 3$\times$3 convolutional layers, but RotateConv kernels with interpolation based angle only can achieve $4/9$ compression ratio on 3$\times$3 convolutional kernels. Considering different emphases on the model size and performance, in this section, we further propose a more efficient interpolation approach aiming at higher compression ratio, i.e., RotateConv based Arithmetic Interpolation(AIRotateConv).

\subsection{Formulation of RotateConv based Arithmetic Interpolation}
Different from RotateConv based angle interpolation which has an additional parameter, i.e., angle, for each convolutional kernel, RotateConv based arithmetic interpolation treats the weights of ``principal components" of a convolution layer as an arithmetic progression, which only requires a tolerance for each convolution layer to do the interpolation calculation. Here, we redefine RotateConv based arithmetic interpolation as
\begin{equation}\small
\begin{aligned}
& C=\left \{\: w,\:  \tau, \: \boldsymbol{\nu}_{i,j} \in V \mid 0\leq i < m ,\: 0\leq j < n \:\right \}\\
& \boldsymbol{\nu}_{i,j}=\left \{v_r \in K_{i,j} \mid 0<r \leq k  \right \}\\
\end{aligned}
\label{eq1}
\end{equation}
where $C$ denotes the parameter set of a pruned convolution layer. $w$ is the minimum weight in reserved weights and $\tau$ is the estimated tolerance of arithmetic progression. The reserved weights of pruned convolution layer are treated as an arithmetic progression. $\boldsymbol{\nu}_{i,j}$ is the point set indicating the ``principal components" of $K_{i,j}$, i.e., the positions of top-$k$ biggest weights of $K_{i,j}$. $V$ is an ordered point set which are ordered by their associated weights, and their associated weights are ranged into an arithmetic progression. $K_{i,j}$ is the $j$-th convolution kernel belongs $i$-th channel, and $k$ is the number of weights of a convolution kernel. 


\subsection{Interpolation based on arithmetic interpolation}\label{m0}
Aiming to extracting the ``principal components'' from regular convolutional kernels, we can initialize the shape of pruned convolutional kernel $\widetilde{K}_{i,j}$ with the ``principal components'' of regular convolutional kernel $K_{i,j}$. Specifically, we sort the weights of $K_{i,j}$ according to their absolute values, then select top-$k$ positions that their absolute weights are bigger than the rest to be the $\boldsymbol{\nu}_{i,j}$.

Similar to RotateConv based angle interpolation, we utilize an algorithm of interpolation to allocate the weights indicated by $V$. Here, we first gather all top-$k$ weights from each $\widetilde{K}_{i,j}$, and sort the them to generate ordered points set $V$. Then, we recalculate weights of points in $V$ through arithmetic interpolation. Specifically, due to the attribute of arithmetic progression, the weights of each $\widetilde{K}_{i,j}$ can be interpolated as
\begin{equation}\small
\begin{aligned}
&w_{v_{i,j,r}}=w+\tau\cdot f(v_{i,j,r})
\end{aligned}
\label{eq8}
\end{equation}
where $v_{i,j,r}$ indicates the $r$-th point in pruned kernel $\widetilde{K}_{i,j}$, and $f(v_{i,j,r})$ denotes an index function which returns the order of point $v_{i,j,r}$ in ordered set $V$. $\tau$ is the estimated tolerance of arithmetic progression which is made up of reserved weights indicating by $V$. Therefore, the estimated tolerance $\tau$ can be calculated by
\begin{equation}\small
\begin{aligned}
&\tau =\frac{\sum_{i=1}^{n}w_{v_{i}}-w_{v_{i-1}}}{n-1},
\end{aligned}
\label{eq9}
\end{equation}
where $n$ is the number of points in $V$. $w_{v_{i}}$ denotes the weight of point $v_i$, and $v_i \in V$.
\subsection{Learning}
Given a pruned convolution layer $C$, there are three kinds of variables need to be learn, i.e., the reserved point set $\boldsymbol{\nu}_{i,j}$ for each regular kernel $K_{i,j}$, the minimum weight $w$ in the reserved points and the tolerance $\tau$. Aiming to learning $\boldsymbol{\nu}_{i,j}$ from regular convolution kernel $K_{i,j}$, we jointly train the network weights of $K_{i,j}$ with sparsity regularization imposed on the latter. Then, we iteratively select kernels with thresholds and top-$k$ points of these reserved kernels which absolute weights are bigger than the rest to be the $\boldsymbol{\nu}_{i,j}$, and recalculate weights of points in $V$ through arithmetic interpolation. Specifically, the training objective of our approach is given by
\begin{equation}\small
\begin{aligned}
&L=\sum_{x,y}l(f(x,W),y)+\lambda \sum_{\gamma \in \Gamma}g(W_{\gamma} )
\end{aligned}
\label{eq10}
\end{equation}
where $(x,y)$ denotes the trained input and target, $W$ denotes the trainable weights of regular convolution, the first sum-term corresponds to the normal training loss of a CNN. $\Gamma$ is the set of layers which will be pruned, $g(W_{\gamma} )$ is a sparsity-induced penalty on the weights of a pruned convolution layer, and $\lambda$ balances the two terms. In this work, we choose $g(W_{\gamma}) = \mid W_{\gamma} \mid$, which is known as $L1$-norm and widely used to feature selection.

Specifically, as shown in Algorithm \ref{learning_algor}, we first normally train the source network in each iteration, and then update $\widetilde{K}$ based on the last $K$ with the methods mentioned in subsection \ref{m0}. Notice that for realizing training our pruned network on existing training tools(e.g., caffe, pytorch), at the end of each iteration, we first set all weights of each $K_{l,i,j}$ to be $0$, then update reserved weights with the weights of $\widetilde{K}_{l,i,j}$.
\begin{algorithm}[h]\small
\caption{Learning RotateConv based Arithmetic Interpolation}
\label{learning_algor}
\begin{algorithmic}[1]
\REQUIRE ~~\\
~~A minibatch of inputs, targets and thresholds $(X, Y, T)$,\\
~~cost function $L(Y,  \widetilde{Y})$,\\
~~a regular model $W=\{K_{l,i,j} \mid 0\leq l <L,\: 0\leq i < m_l ,\: 0\leq j < n_l\}$;\\ 
\ENSURE~~\\
~~$\Phi :=\{w_l,\tau_l,\widetilde{K}_{l,i,j} \mid l \in \Gamma,\: 0\leq i < m_l ,\: 0\leq j < n_l \}$;\\
\REPEAT
\STATE $\widetilde{Y}$=Forward $(X,W)$;
\STATE $\frac{\partial L}{\partial W}$=Backward $(\frac{\partial L}{\partial \widetilde{Y}})$;
\STATE $W^{t+1}$=UpdateParameters $(W^{t},\frac{\partial L}{\partial W})$;
\FOR{$l \in \Gamma$}
\FOR{each $K_{l,i,j}$}
\IF{$max(|K_{l,i,j}|)< t_l, t_l \in T$}
\STATE continue;
\ENDIF
\STATE sort points in $K_{l,i,j}$ according to their absolute weights;
\STATE select top-$k$ points to be the set $\boldsymbol{\nu}_{l,i,j}$;
\ENDFOR
\STATE gather $\boldsymbol{\nu}_{l,*,*}$ into $V_l$;
\STATE sort points in $V_l$ according to their weights;
\STATE calculate estimated tolerance $\tau_l$ with Eq.(\ref{eq9});
\STATE interpolate values to the points in $V_l$ with Eq.(\ref{eq8});
\FOR{each $\widetilde{K}_{l,i,j}$}
\STATE update the weights of points in $\boldsymbol{\nu}_{l,i,j}$;
\STATE update the weights of $K_{l,i,j}$ with the weights of $\widetilde{K}_{l,i,j}$;
\ENDFOR
\ENDFOR
\UNTIL{convergence}
\end{algorithmic}
\end{algorithm}

\section{Experiment}
For evaluating the effectiveness of proposed methods, we study the performance of pruned models generated by proposed methods on the tasks of classification and objective detection. In this section, we first introduce datasets, base-lined models and experimental settings respectively. Then, we evaluate performance of RotateConv and AIRotateConv models which do interpolation based on angle and arithmetic progression, respectively.

\subsection{Datasets}
CIFAR\cite{krizhevsky2009cifar} consists of colored natural images with 32$\times$32 pixels. CIFAR-10 consists of images drawn from 10 classes and CIFAR-100 from 100 classes. The training and testing sets contain 50,000 and 10,000 images respectively, and hold out 5,000 training images as a validation set. In our experiments, the input data is normalized using channel means and standard deviations without any data augmentation. For the final run we report the final test error on the test set at the end of training.

SVHN\cite{netzer2011reading} is a real-world digit image dataset for developing machine learning and object recognition algorithms. It is obtained from house numbers in Google Street View images. The task is to classify the digit centered in image. It has 10 classes, 73,257 digits for training, 26,032 digits for testing. We use the CIFAR-like version for experiments, each image has a ${32\times32}$ spatial size and is centered around a single digit which means that many examples do contain some distractions at the sides.

PASCAL VOC\cite{Everingham10} is a real-world digit image dataset for developing object detection and localization algorithms. Our training dataset consists of a set of images from the training datasets of VOC-2007 and VOC-2012, and the testing dataset(4952 images) is the testing set of VOC-2007. Each image has an annotation file giving a bounding box and object class label for each object in one of the twenty classes present in the image. We adopt the same configurations with SSD\cite{liu2016ssd} in training and testing, and report mAP performance of SSD300(the size of input images is 300$\times$300 pixels).

\subsection{Deep models}
For studying the performance of our methods both on light and heavy models, we select three kinds of network structures, i.e., ResNet\cite{he2016deep}, DenseNet\cite{huang2016densely} and VGG-Net\cite{vgg} to do experiments.

ResNet is a kind of popular network structure in modern CNNs and makes great contribution in deep learning. Using shortcut connections and deeper networks, it massively improves the performance in various learning tasks while maintaining the efficiency in the model size.

DenseNets are improved from ResNet. They connect each layer to every other layer in a feed-forward fashion. In this way, DenseNets alleviate the vanishing-gradient problem, strengthen feature propagation, encourage feature reuse, and substantially reduce the number of parameters. In our experiments, we reproduce a 40-layer DenseNet as illustrated in \cite{he2016deep} with growth rate 60.

VGG-Net is a neural network that performed very well in the Image Net Large Scale Visual Recognition Challenge (ILSVRC) in 2014. It scored first place on the image localization task and second place on the image classification task. Only 3$\times$3 convolution and 2$\times$2 pooling are used throughout the whole network. VGG also shows that the depth of the network plays an important role and deeper networks give better results. In our experiments, we also adopt the SSD300 architecture\cite{liu2016ssd} to do experiments for evaluating performance of our methods on the task of object detection. SSD300 utilizes VGG-16\cite{vgg} as its base architecture, which only removes all the dropout layers and the fc8 layer. For obtaining predictions of detections at multiple scales, SSD300 adds convolutional feature layers which decrease in size progressively to the end of the truncated base network.

\subsection{Experimental setting}
All the network structures are trained using stochastic gradient descent (SGD). On CIFAR, we train the baselines, i.e., ResNet and DenseNet, using batch size 64 for 300 epochs without data augmentation. The initial learning rate is set to 0.01, is divided by 10 at 50\% and 75\% of the total number of training epochs. We use a weight decay of $10^{-4}$ and a Nesterov momentum\cite{momentum} of 0.9 without dampening. The weight initialization introduced by \cite{xviar} is adopted. On PASCAL VOC, we report mAP of the SSD300 model provided by the authors. The backbone of SSD300 is VGG-16 which is pre-trained on the ILSVRC CLS-LOC dataset\cite{ilcvrcDataset} with initial learning rate 0.001, 0.9 momentum, 0.0005 weight decay, and batch size 32.

We train pruned networks which 3$\times$3 convolution kernels are set to be pruned. For each pruned convolution layer, a Batch Normalization\cite{bn} layer is added after that. On CIFAR, we train prunned ResNet and DenseNet with the initial learning rate $0.01$, and keeping other settings the same as their baselines. On PASCAL VOC, we train the pruned SSD300 following the same settings in baseline training. 

\subsection{Results and analysis}
In this section, we first analyze the performance of RotateConv models which are generated by the method mentioned in section \ref{rck}, as well as the improved models which are produced by the approach of AIRotateConv mentioned in section \ref{ai}. Then, we observe the effect of applying AIRotateConv on different layers and the number of pruned layers to explore a appropriate pruning rule.


\subsubsection{Effectiveness of approach}
The numbers of parameters of deep neural networks can be reduce a lot by our proposed approaches as well as their performance still can be acceptable. As shown in Table~\ref{my-label}, Version-$9$ means the basic network which most of convolution layers are ${3\times3}$ convolution, Version-$4$ and Version-$3$ means 4 parameters and 3 parameters models pruned by RotateConv, while Version-$3^{*}$ means 3 parameters model pruned by AIRotateConv with threshold $0.001$, respectively. Taking ResNet20 for an example, the performance of the various versions is similar, while the basic model has ${16.62M}$ parameters, the pruned models, i.e., Version-$4$, Version-$3$ and Version-$3^{*}$ models only have ${9.23M}$, ${5.54M}$ and ${4.71M}$ parameters, respectively. Moreover, according to the experimental results shown in Table~\ref{my-label}, the performance of Version-$4$ is even better than Version-$9$ in some cases. It indicates that the line segment kernels still have a powerful capability on feature extraction, and sometimes even have better generalization than traditional square kernels, because they have deformable kernels.
\begin{table}[]
\centering
\caption{\small Performance of image classification on CIFAR-10/100 and SVHN datasets. ``\#Params'' denotes the number of convolutional parameters in the model. Three settings are listed for comparison, as shown in second row on table, 9 for ${3\times3}$ baseline kernel, 4 for 4 parameters RotateConv version, 3 for 3 parameters RotateConv version, and $3^{*}$ for 3 parameters AIRotateConv version which cutting threshold is $0.001$ . All 3$\times$3 $conv$ layers in each network are pruned, except their first layers.} \vspace{0.5em}
\label{my-label}
\scriptsize
\begin{tabular}{lccccccccccccccc}
\hline
\multicolumn{1}{|l|}{Model}     & \multicolumn{4}{c|}{ResNet20}              & \multicolumn{4}{c|}{VGG}              & \multicolumn{4}{c|}{DenseNet40}            \\ \hline

\multicolumn{1}{|l|}{Version}   & 9\cite{he2016deep}    & 4 & 3 & \multicolumn{1}{c|}{$3^{*}$}   & 9\cite{he2016deep}     & 4   & 3  & \multicolumn{1}{c|}{$3^{*}$} & 9\cite{huang2016densely}   & 4  & 3 & \multicolumn{1}{c|}{$3^{*}$}   \\

\multicolumn{1}{|l|}{\#Params}   & 16.62M  & 9.23M &5.54M & \multicolumn{1}{c|}{4.71M}   & 19.09M  & 10.61M & 6.36M & \multicolumn{1}{c|}{5.63M}  & 4.03M  & 2.24M &1.35M& \multicolumn{1}{c|}{1.22M}\\ \hline


\multicolumn{1}{|l|}{CIFAR-10}      & \textbf{91.70}  & 90.78 & 84.83 & \multicolumn{1}{c|}{90.28}    & \textbf{92.52}  & 91.83  & 83.35 & \multicolumn{1}{c|}{91.06} & 92.63 & \textbf{92.68} &81.24 & \multicolumn{1}{c|}{90.89}  \\

\multicolumn{1}{|l|}{CIFAR-100}     & 52.84 & \textbf{53.45}& 50.39& \multicolumn{1}{c|}{52.75} & 56.63 & \textbf{56.93}& 51.32& \multicolumn{1}{c|}{56.12}  & \textbf{72.45} & 69.99&50.64 &\multicolumn{1}{c|}{68.64}\\

\multicolumn{1}{|l|}{SVHN}   & 95.82 & \textbf{96.01} & 95.17& \multicolumn{1}{c|}{95.69} & 96.02  & \textbf{96.40} & 96.19& \multicolumn{1}{c|}{95.68} & \textbf{97.15} &  96.84  &  90.69&  \multicolumn{1}{c|}{96.79}\\ \hline
\end{tabular}
\end{table}

However, the experimental results in Table~\ref{my-label} also indicate that the performance of Version-$3$ and Version-$4$ exist a degree of gap. This phenomenon shows that weights diversity is helpful for feature extraction and has a significant impact on performance. For example, Version-$4$ can directly model a triangle with a line segment kernel, but Version-$3$ can not. Thus, the performance of Version-$3$ usually decline more than Version-$4$. In addition, the performance of Version-$3^{*}$ can be compared to Version-$4$ while its parameters are even less than Version-$3$. This result is because AIRotateConv not only can learn the key positions in convolution kernels by $L1$-norm and thresholds, but also utilize arithmetic interpolation to the full extent of maintaining weights diversity.
\begin{figure*}[htbp]
\centering
 \subfigure[ Performance of training models from scratch]
{\includegraphics[height=0.35\linewidth,width=0.47\linewidth]{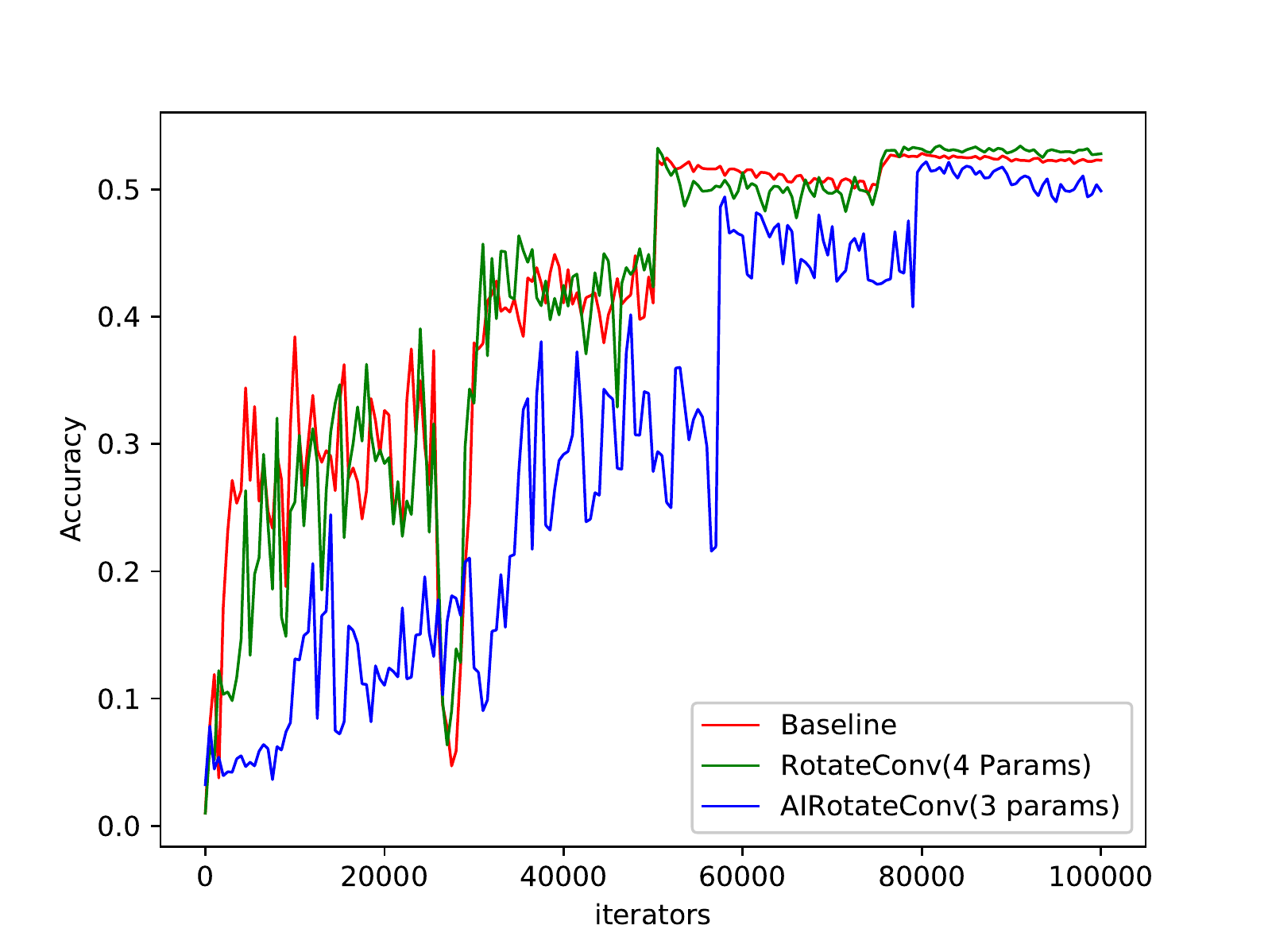}}
 \subfigure[ Performance of fine-tuning pruned models]
{\includegraphics[height=0.35\linewidth,width=0.47\linewidth]{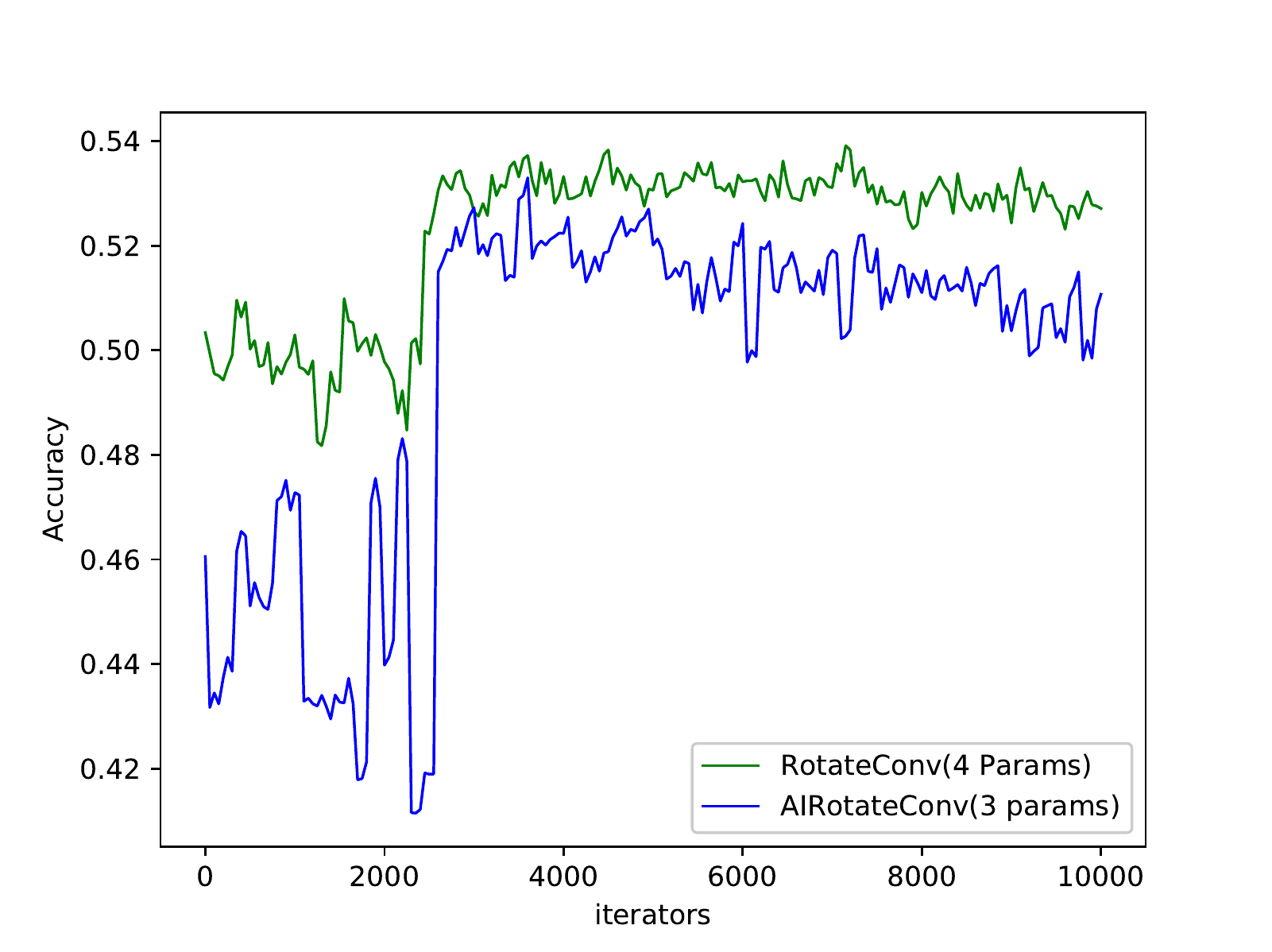}}
\caption{Convergence of ResNet20 on CIFAR100.} 
\label{fig4}
\end{figure*}

Figure~\ref{fig4} shows the accuracy curves for various versions of ResNet20 on the dataset CIFAR-100. As we can see in the Figure~\ref{fig4}(a), accuracies of Baseline, RotateConv and AIRotateConv seesaw by iterations, and finally get almost equal performance. The ascent tendency of AIRotateConv is a little slower than others when it was trained from scratch. This phenomenon may demonstrate that the rate of convergence of AIRotateConv is more relied on parameter initialization, because it needs to select key positions of convolutional kernels. Therefore, we usually initialize the pruned models of AIRotateConv by trained basic models, and then fine-tune the pruned models with relatively small learning rate, e.g., 0.001. In this way, we can quickly obtain an acceptable pruned model by AIRotateConv shown in Figure~\ref{fig4}(b).

\begin{table}[]
\centering
\caption{\small Test Errors of pruned DenseNet-40 models on CIFAR-10.  NS(40\% Pruned) and NS(70\% Pruned) denotes the models which 40\% and 70\% convolutional channels are pruned by Network Slimming\cite{learningEfficient}, respectively. Version-4 denotes the model pruned by RotateConv with 4 parameters, Version-$3^{*}$ is the model pruned by AIRotateConv with 3 parameters and 0.001 threshold.} \vspace{0.5em}
\label{comTab}
\scriptsize
\begin{tabular}{|c|ccccccc}
\hline
Model
     & err(\%)       &  pruned params     & \multicolumn{1}{c|}{pruned flops}   \\\hline


\multicolumn{1}{|l|}{Baseline}           & 7.37   &  --        &\multicolumn{1}{c|}{--}   \\

\multicolumn{1}{|l|}{NS(40\% Pruned)}    & 6.11    & 35.7\%    &\multicolumn{1}{c|}{28.4\%} \\

\multicolumn{1}{|l|}{NS(70\% Pruned)}    & 5.19    & 65.2\%    &\multicolumn{1}{c|}{55.0\%} \\

\multicolumn{1}{|l|}{version-4}          & 7.32     &44.35\%   &\multicolumn{1}{c|}{38.15\%}  \\

\multicolumn{1}{|l|}{version-$3^{*}$}    & 9.11     &69.72\%    &\multicolumn{1}{c|}{59.67\%}\\ \hline
\end{tabular}
\end{table}

One purpose of this work is to reduce the amount of computing resources needed. From Table~\ref{comTab}, we can observe that, on DenseNet, typically when 44\%-69\% convolution parameters are pruned, our aproaches, i.e.,version-4 and version-$3^{*}$ can achieve acceptable performance comparing with the original models and other pruning methods. For example, when  69\% convolution parameters and 59.67\% flops are pruned, version-$3^{*}$ still can achieve a test error of 9.11\% on CIFAR-10. Although our methods are inferior on the viewpoint of test errors, they can achieve higher compression ratio and less computation than the slimming model\cite{learningEfficient}.
\begin{figure*}[htbp]\scriptsize
\setlength{\abovecaptionskip}{3pt}
\setlength{\belowcaptionskip}{-7pt}
\centering
 \subfigure[Kernel angles along input channels]{\includegraphics[height=0.33\textwidth, width=0.46\textwidth]{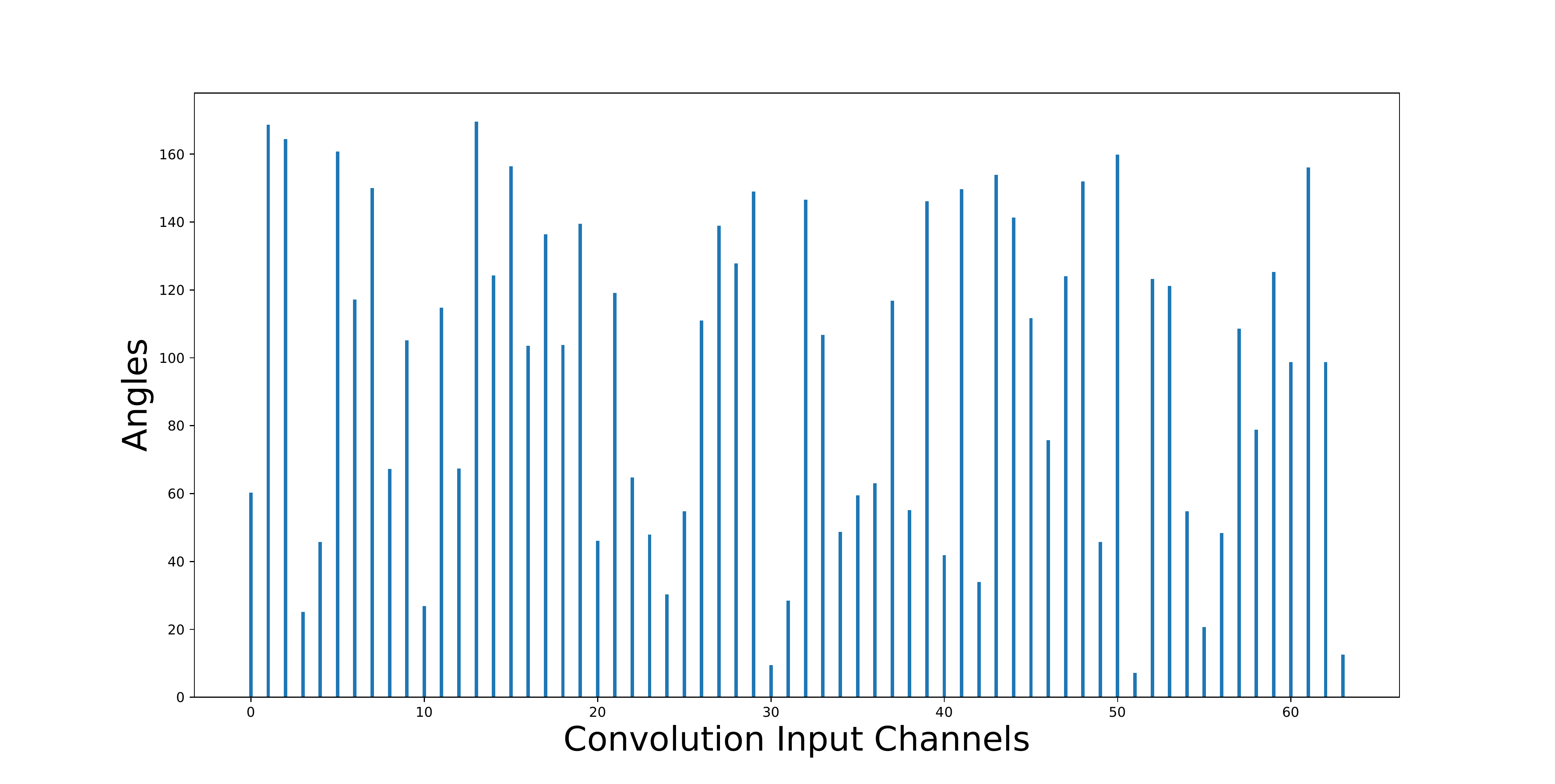}}\label{aa}
 \hspace{1.5ex}
 \subfigure[Kernel angles along output channels]
 {\includegraphics[height=0.33\textwidth, width=0.46\textwidth]{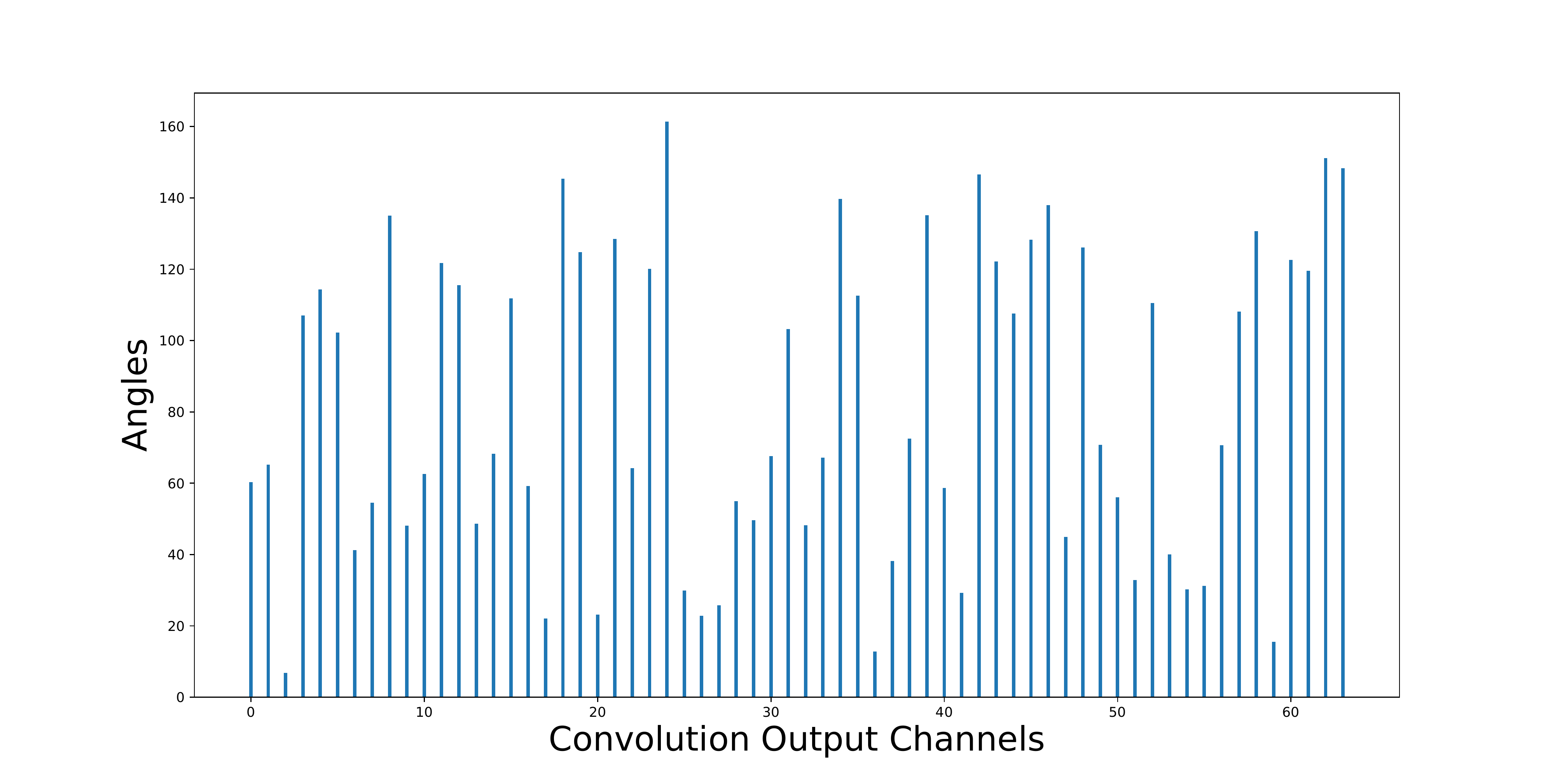}}\label{bb}
 \caption{ Kernel angle distributions for layer $Conv$-$21$ in ${ResNet20}$ trained on CIFAR-100. Layer $Conv$-$21$ has 64 input channels and 64 output channels. (a) For the first output channel, the 64 kernel angles applied on input features. (b) For the first input channel, the 64 kernel angles used by 64 output channels respectively.}
 \label{fig6}
\end{figure*}

\subsubsection{Analysis of working mechanism}
For detecting the working mechanism of line segment kernels, we choose the model ${ResNet20}$ without loss of generality, and train it on CIFAR-100 with 4 parameters RotateConv kernels. Then, we observe the angle distributions of last convolutional layer $Conv$-$21$, which has 64 output channels and 64 input channels. Here, we analyze two distributions along input and output channel respectively.

On the one hand, we give the angle distribution for one output channel. For each output channel, it has 64 spatial kernels corresponding to 64 RotateConv angles applied along input channels. Figure~\ref{fig6}(a) shows the angle distribution of the first output channel, which is denoted as ${\{{\theta _{1,j}}|j = 1,2,..,64\}}$ before. We can find that different input features are applied with different RotateConv angles, which is coincident with universal intuition that different features represent different patterns.

On the other hand, we give the angle distribution for one input channel. For each input channel, it has been repeatedly used by 64 different output channels which has 64 different RotateConv angles too. Figure~\ref{fig6}(b) shows the angle distribution of the first input channel, which is denoted as ${\{{\theta _{i,1}}|i = 1,2,..,64\}}$ before. We can find that one single input feature is repeatedly applied with different RotateConv angles, which can be explained that one feature map always contains various patterns and the later operations need respectively select these patterns for further processing.
\begin{figure*}[htbp]\scriptsize
\setlength{\abovecaptionskip}{3pt}
\setlength{\belowcaptionskip}{-7pt}
\centering
 \subfigure[ ]
 {\includegraphics[width=0.48\textwidth]{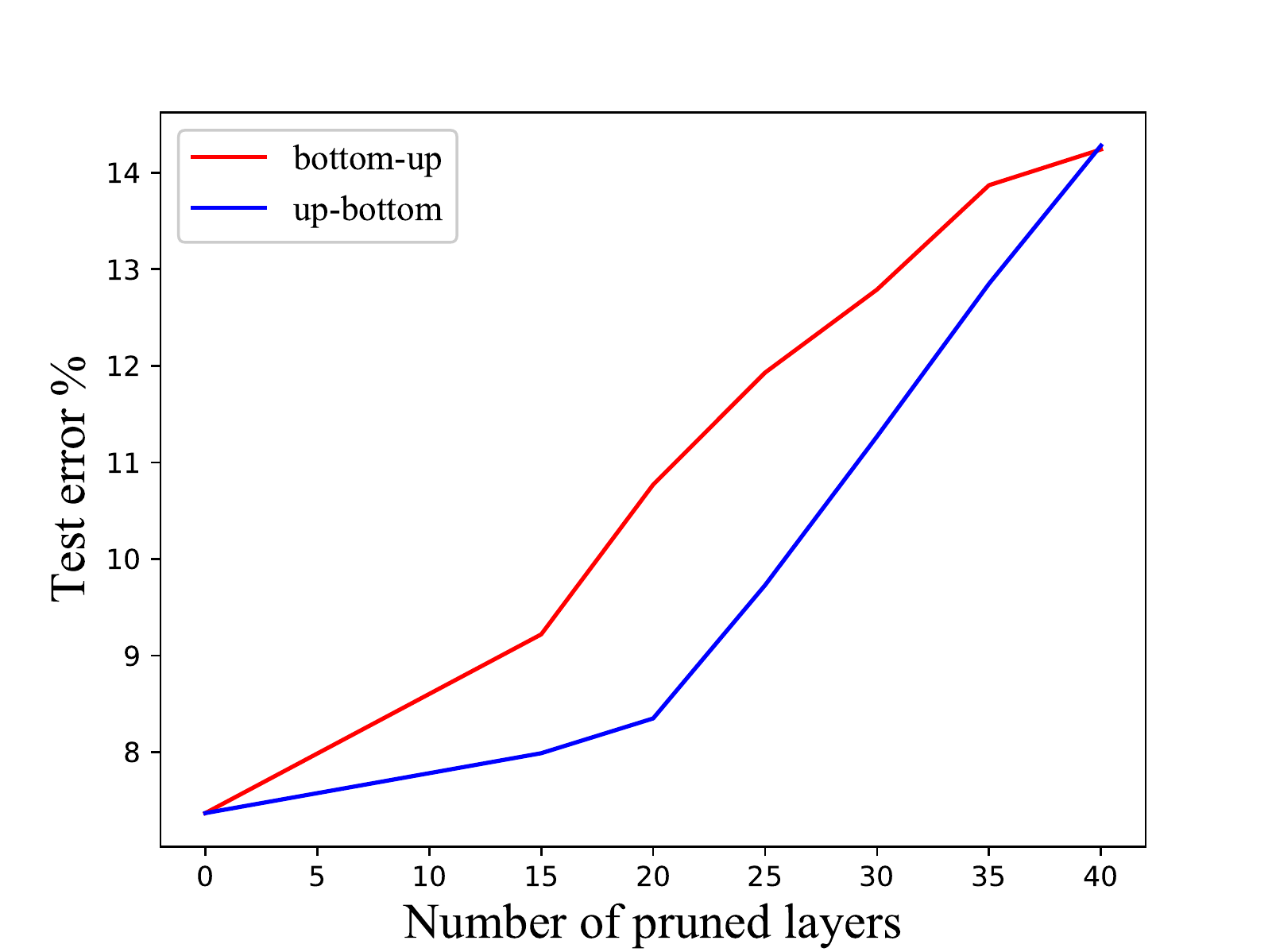}}\label{a}
 \hspace{1.5ex}
 \subfigure[ ]
 {\includegraphics[width=0.45\textwidth]{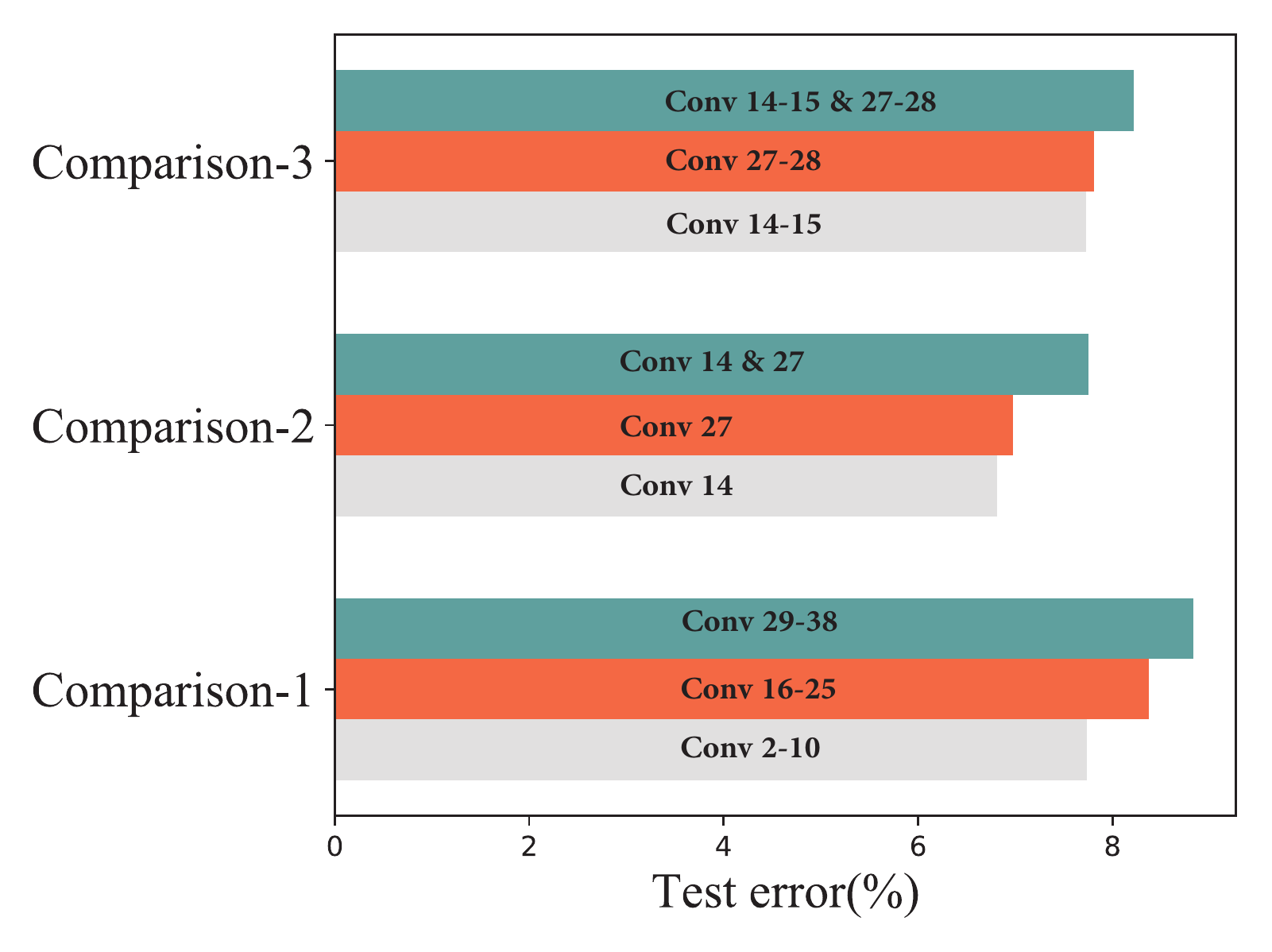}}\label{b}
 \caption{ Effect of pruning different layers with AIRotateConv. (a) Accuracy varying with the increasing number of pruned layers. (b) Effect of independently pruning different parts of network.}
 \label{effect}
\end{figure*}
\subsubsection{Effect of pruning different layers}
To study the effect of pruning different layers, we observe the accuracy of image classification varying pruning number of layers by AIRotateConv. Without loss of generality, we prune DenseNet-$40$ from bottom-to-up and up-to-bottom, respectively. As shown in Figure~\ref{effect}(a), the accuracies usually decline with the number of pruned layers increasing, but the direction of bottom-to-up declines more quickly than the direction of up-to-bottom at initial stage. For example, at the stage of pruning $0$ to $10$ layers, the performance degradation of bottom-to-up is significantly faster than up-to-bottom. It indicates that the low-level features seemingly require relatively powerful convolutional kernels to do feature extraction, because the low-level features usually are detailed information. Therefore, in practice, we usually maintain the first few layers without pruning for balancing the accuracy and the resources reductions. Besides, we can observe from Figure~\ref{effect}(b) that the high layers seem to be more sensitive to be pruned than the low layers. For example, when the low(conv2 to conv10), middle(conv16 to conv25), and high(conv29 to conv38) layers are pruned respectively, the lower pruning cases seem to outperform higher cases. This is due to the high layers usually process abstract information which is crucial to classification. Therefore, according to requirements of balancing the accuracy and the resources reductions, we usually maintain the first and latest convolution layers without pruning.
\begin{table}
\centering
\caption{\small Performance of AIRotateConv approach on SSD300 with varying pruning ratio. Baseline indicates the original SSD300 without any pruning. Model-1 is the pruned SSD300 which all convolution layers are applied pruning, while Model-2 is the pruned SSD300 maintaining the first $Conv$ and latest layer without pruned. Model-3, Model-4 and Model-5 are the pruned SSD300 which the first and latest $Conv$ layer are not pruned, and their insignificant kernels are cast off with different thresholds.}\vspace{0.5em}
\label{Tab3}
\scriptsize
\begin{tabular}{|lccccccccc}
\hline
\multicolumn{2}{|c}{Model}&mAP &\multicolumn{2}{c}{Threshold} & \multicolumn{2}{c}{Param pruned}& \multicolumn{2}{c|}{FLOP pruned} \\
\hline
\hline
\multicolumn{2}{|c}{Baseline}&74.0 & \multicolumn{2}{c}{--} &\multicolumn{2}{c}{--} &\multicolumn{2}{c|}{--} \\
\multicolumn{2}{|c}{Model-1}&67.1 & \multicolumn{2}{c}{0.001} &\multicolumn{2}{c}{74.7\%} &\multicolumn{2}{c|}{79.3\%} \\
\multicolumn{2}{|c}{Model-2}&73.8 & \multicolumn{2}{c}{0.001} &\multicolumn{2}{c}{74.7\%} &\multicolumn{2}{c|}{79.1\%} \\

\multicolumn{2}{|c}{Model-3}&73.2 & \multicolumn{2}{c}{0.0015} &\multicolumn{2}{c}{77.2\%} &\multicolumn{2}{c|}{83.1\%} \\
\multicolumn{2}{|c}{Model-4}&70.8 & \multicolumn{2}{c}{0.0020} &\multicolumn{2}{c}{79.5\%} &\multicolumn{2}{c|}{83.7\%} \\
\multicolumn{2}{|c}{Model-5}&66.7 & \multicolumn{2}{c}{0.0035} &\multicolumn{2}{c}{86.3\%} &\multicolumn{2}{c|}{86.2\%}\\
\hline
\end{tabular}
\end{table}

Keeping the observation of Figure~\ref{effect} in mind, we further study the performance of AIRotateConv approach applying on the object detection network, i.e., SSD300. As shown in Table~\ref{Tab3}, applying AIRotateConv approach on SSD300 can reduce amount of parameters while the performance of pruned models are acceptable. Moreover, due to maintaining the first convolution layer without punning, the mAP of Model-2 outperforms Model-1. This result reconfirms that the low-level features seemingly require relatively powerful convolutional kernels, and we usually keep the first convolution layer without pruning in practice for balancing the accuracy and the resources reductions. For obtaining more condensed models, we further remove the kernels which absolute $\rho$ are smaller than given thresholds, i.e., ignore the associated input channels for each layer when their weights of convolutional kernels are close to 0. The experimental results are shown by Model-3, Model-4 and Model-5. It indicates that as the threshold increasing, the parameter reduction ratio is increasing and the performance indicated by mAP still can be acceptable.

\section{Conclusion}
The aim of this work is to reduce computing resource requirements of CNNs as well as maintain their performance. Thus, we propose a kind of convolutional kernel which has extremely simple shape as line segments, and equip them with the rotatable ability to model diverse features. The rotatable ability is achieved by using inverse-interpolation which makes angles continuous, differentiable and learnable. In this paper, we use RotateConv and AIRotateConv to significantly reduce the number of model parameters, as well as maintain the accuracies of models. The difference between these variants is the method of interpolation, i.e., interpolation of RotateConv is based on angles while AIRotateConv does inverse-interpolation with arithmetic interpolation. In experiments, three kinds of network structures, i.e., ResNet20, VGG and DenseNet40 are pruned for efficiency analysis and pruning strategies exploration.

In the future, we will devote to the following problems. Firstly, proposed approaches should be validated on more large scale datasets such as ImageNet\cite{russakovsky2015imagenet} and COCO\cite{lin2014microsoft}. Secondly, RotateConv is achieved by inverse-interpolation with angle parameter, but it brings RotateConv kernels back to original shape for computation, and does not reduce computation. Therefore, we hope more efforts could be devoted to make further progress in acceleration of RotateConv. Last, although AIRotateConv has the potential to reduce the number of FLOPs and make an acceleration, it still needs to modify existing software framework like caffe, meanwhile, FPGAs also are main platforms for model inference, we plan to involve AIRotateConv into high-performance frameworks, e.g., ncnn, for low-power CPUs and FPGAs.

\bibliographystyle{unsrt}
\bibliography{sigproc}

\end{document}